\documentclass[conference]{IEEEtran}
\IEEEoverridecommandlockouts
\usepackage{cite}
\usepackage{amsmath,amssymb,amsfonts}
\usepackage{algorithmic}
\usepackage{graphicx}
\usepackage{textcomp}
\usepackage{xcolor}
\usepackage{hyperref}
\usepackage{arydshln}
\usepackage{subcaption}
\usepackage{multicol, multirow}
\usepackage{float}
\usepackage{pifont}
\usepackage[font=footnotesize]{caption}
\def\BibTeX{{\rm B\kern-.05em{\sc i\kern-.025em b}\kern-.08em`
    T\kern-.1667em\lower.7ex\hbox{E}\kern-.125emX}}
\usepackage{pifont}
\usepackage{xcolor}
\usepackage{caption}
\usepackage{listings}
\usepackage{xcolor}
\usepackage{booktabs}
\usepackage{arydshln}
\usepackage{pgfplots}
\pgfplotsset{compat=1.18}
\usepackage{tikz}
\usepackage{rotating}
\usepackage{dblfloatfix}

\usepackage{todonotes}
\usepackage{booktabs}
\title{\bf
Towards Scaling Marine Perception with Synthetic Data
}

\author{Haoyu Ma$^{1}$, Onur Bagoren$^{1}$, Anja Sheppard$^{1}$,  Elias Fandi$^{1}$, \\ Ashrith Edukulla$^{1}$, Tanner Aslan$^{2}$, Natasha Sieh$^{1}$, Jingyu Song$^{1}$, and Katherine A. Skinner$^{1}$
}

\begin{document}
\twocolumn[{
    \renewcommand\twocolumn[1][]{#1}
    \maketitle
    \centering
\includegraphics[width=\linewidth]{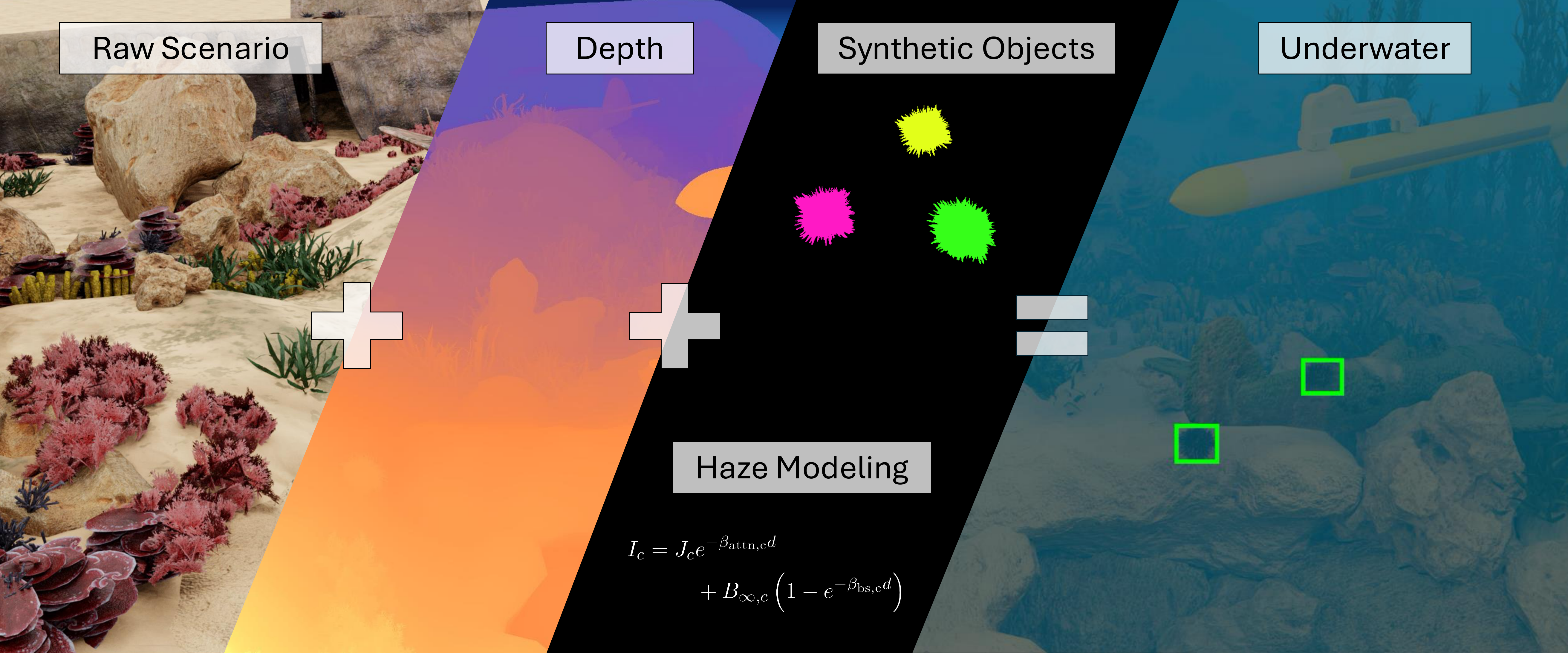}
    \captionof{figure}{OceanSim rendering pipeline overview, with the additional ability to add labeled synthetic objects with our proposed Synthetic Data Generation pipeline.}
    \label{fig:overview}
    \vspace{1em}
}]

\begin{NoHyper}
\begingroup
\renewcommand\thefootnote{}%
\footnote{This work was supported by National Science Foundation Grant \#DGE-2241144.}
\footnote{$^{1}$ Department of Robotics, University of Michigan, Ann Arbor, MI, USA}
\footnote{$^{2}$ Department of Computer Science and Engineering, University of Michigan, Ann Arbor, MI, USA}
\footnote{Corresponding author e-mail: {\tt\small obagoren@umich.edu}}
\addtocounter{footnote}{-1}%
\endgroup
\end{NoHyper}

\begin{abstract}

Scalable machine learning in challenging underwater environments is strongly limited by the lack of labeled real-world training data.
This data is often expensive and laborious to gather, making large-scale real-world data challenging to gather and curate.
However, simulated data can help close the gap, enabling many learning-based tasks for underwater perception.
In this work, we extend OceanSim, an IsaacSim-based underwater perception simulator, with a Synthetic Data Generation (SDG) pipeline for training models to be used in underwater scenarios.
The proposed pipeline enables users to generate large, automatically labeled, photorealistic datasets with configurable scene appearance, structure, and sensor settings.
We evaluate the pipeline on a real-world sea urchin detection task and study how different forms of synthetic scene variation affect sim-to-real performance.
Based on these experiments, we discuss findings on our results, main limitations of the current pipeline and identify future directions for improving underwater rendering fidelity, scene diversity, and the evaluation of sim-to-real generalization.
The open-source code can be found at \url{https://github.com/umfieldrobotics/OceanSim}.
\end{abstract}
\section{Introduction}

Robust perception in underwater robotics remains fundamentally constrained by data scarcity and severe degradation induced by the environment~\cite{jian2024uoddatasetsurvey}. 
The collection of data for improving underwater perception is expensive and labor-intensive, limiting the development of data-hungry end-to-end learning methods for tasks such as object detection, semantic segmentation and mapping.

\begin{table*}[t]
    \centering
    \renewcommand{\arraystretch}{1.5} 
    \caption{Comparison of existing underwater simulation frameworks in terms of simulation engine, annotation and synthetic data generation (SDG) support, ROS integration, and supported sensors. DVL denotes a Doppler velocity log, FLS denotes a forward-looking-sonar, and SSS denotes a side-scan sonar.}
    \begin{tabular}{l|ccccccccc}
        \toprule
        
        \multirow{2}{*}{\textbf{Simulator}} &
        \multirow{2}{*}{\textbf{Engine}} &
        \multirow{2}{*}{\textbf{Annotations}} &
        \multirow{2}{*}{\textbf{SDG Support}} &
        \multirow{2}{*}{\textbf{ROS Support}} &
        \multicolumn{5}{c}{\textbf{Sensor Support}} \\
        
        & & & & &
        \textbf{RGB-D} &
        \textbf{DVL} &
        \textbf{Barometer} &
        \textbf{FLS} &
        \textbf{SSS} \\
        
        \midrule
        \midrule
        
        Stonefish~\cite{cieslak2019stonefish, grimaldi2025stonefish}
        & Custom/OpenGL
        & \checkmark
        &
        & ROS1 \& ROS2
        & \checkmark
        & \checkmark
        &
        & \checkmark
        & \\
        
        HoloOcean~\cite{potokar2022holoocean, potokar2022holooceansonar, holoocean2024, romrell2025preview}
        & Unreal Engine 5
        & \checkmark
        &
        & ROS1 \& ROS2
        & \checkmark
        & \checkmark
        & \checkmark
        & \checkmark
        & \checkmark \\
        
        UNav-Sim~\cite{amer2023unav}
        & Unreal Engine 5
        & 
        & \checkmark
        & ROS1 \& ROS2
        & \checkmark
        &
        &
        &
        & \\

        MarineGym~\cite{chu2025marinegym}
        & IsaacSim 
        & 
        &
        & ROS1
        & 
        &
        &
        &
        & \\

        Aqua-Verse~\cite{bai2026aqua}
        & Unreal Engine 5
        & 
        &
        &  
        & \checkmark 
        & \checkmark 
        & 
        & \checkmark
        & \\

        LOTUSim~\cite{buche2026lotusim} 
        & Gazebo
        & 
        &
        & ROS2 
        & \checkmark 
        & 
        & 
        & 
        & \\
        
        OceanSim~\cite{song2025oceansim}
        & IsaacSim
        & \checkmark
        & \checkmark
        & ROS2
        & \checkmark
        & \checkmark
        & \checkmark
        & \checkmark
        & \\
        
        \bottomrule
    \end{tabular}
    \label{tab:sim_overview}
\end{table*}

The underwater domain is particularly challenging for perception tasks.
Particularly for vision-based systems, the light attenuation, backscatter from suspended particles, and high variability in illumination and turbidity cause appearance distributions to vary significantly across sites, depths, and sensor configurations~\cite{akkaynak2018revised, monzon2024spectral}.
When real-world data is collected, the high variability in visual appearance often causes poor generalization to other sites~\cite{duras2024seaclear, chen2024uodreview}.

Synthetic data generation (SDG) has emerged as a promising solution to overcome the generalizability challenges of underwater vision systems by enabling scalable, controllable, and cost-effective simulated data generation~\cite{tremblay2018training, sethuraman2023stars, grimaldi2025stonefish}. 
Through SDG, datasets used for training models are no longer constrained by site access or robot hardware, making it an ideal solution for open perception problems in the field~\cite{amer2023unav, cavalieri2026physically}. 
However, a key challenge exists in producing realistic imagery that matches the real underwater distribution~\cite{aldhaheri2025underwater}.

Previous work has focused on reducing the sim-to-real gap in underwater domains by improving sensor models and rendered photorealism~\cite{song2025oceansim, amer2023unav, holoocean2024, gomes2025navigating}. 
The OceanSim simulator is an IsaacSim-based platform which focuses on realistic perceptual data rendered at high speeds~\cite{song2025oceansim}. 
Data generated from OceanSim has already enabled perception advancements such as opti-acoustic fusion~\cite{chen2025sonarsweep} and underwater legged locomotion~\cite{aina2026review}. 
However, OceanSim currently lacks a straightforward pipeline for generating large amounts of synthetic data and labels via SDG for learning tasks.


In this work, we explore methods to generate realistic data as an extension to OceanSim for training models for generalization of a common vision task (object detection) and discuss the challenges with sensor realism in underwater simulated environments. 
This pipeline is demonstrated visually in Fig.~\ref{fig:overview}.
The proposed extension of OceanSim includes:
\begin{enumerate}
    \item A highly-configurable OceanSim synthetic data generation pipeline for machine learning applications, 
    \item An OceanSim ROS2 wrapper for easy integration into downstream applications.
\end{enumerate}

Overall, our goal is to enable real-world perception tasks, such as object detection and segmentation, using high-fidelity simulation. 
To assess OceanSim's perceptual fidelity, we perform benchmarking on an underwater object detection task trained entirely in simulation. 
Although we feel that these contributions to OceanSim will help enable real-world perception tasks, we also include an in-depth discussion of open challenges that require further attention from the community.

\section{Related Work}

\subsection{Synthetic Data for Perception}

Synthetic data has long been used as a substitute or complement to real-world annotations in perception pipelines, particularly when labeling is expensive, unsafe, or limited in coverage~\cite{sethuraman2023stars, cavalieri2026physically, johnsonrobertson2017driving}. 
Tremblay et al. systematically analyzed how randomized simulation parameters affect detection performance and showed that synthetic data can serve both as a primary training source and as a pretraining signal for improving performance with limited real data~\cite{tremblay2018training}.

Recent work has focused on scalable SDG pipelines that provide rich supervision, including segmentation masks, depth, bounding boxes, and occlusion metadata. SynTable~\cite{ng2023syntable} demonstrates that photorealistic SDG with structured annotations improves sim-to-real transfer in cluttered tabletop scenarios. Similarly, Syn2Real~\cite{agrawal2024syn2real} and STARS~\cite{sethuraman2023stars} report significant performance gains in side-scan sonar detection tasks when augmenting scarce real data synthetic samples. These results highlight synthetic data as an effective strategy for deployment-oriented perception systems when only limited real data is available. However, underwater environments introduce additional complexities in modeling synthetic data, and naive domain randomization approaches designed for terrestrial settings often fail to achieve successful sim-to-real transfer~\cite{chen2024uodreview}.

\subsection{Underwater Simulators}

Well-designed SDG pipelines can help cross the sim-to-real barrier. In the underwater domain, there are several simulators with varying degrees of support for SDG. Stonefish~\cite{cieslak2019stonefish,grimaldi2025stonefish} is built on a custom OpenGL-based C++ platform, and supports a variety of sensors as well as exporting semantic annotation labels. However, it does not have a dedicated SDG pipeline, making the generation of large datasets challenging. Similarly, HoloOcean~\cite{potokar2022holoocean, potokar2022holooceansonar, holoocean2024, romrell2025preview}, while having strong support for different sensor modalities including SSS, does not have support for scalable SDG. UNav-Sim~\cite{amer2023unav} has a dedicated SDG approach, but lacks support for a variety of sensors. MarineGym~\cite{chu2025marinegym} and LOTUSim~\cite{buche2026lotusim} are focused more on robust underwater dynamics or human-in-the-loop, with little support for physics-based sensors or SDG. Aqua-Verse~\cite{bai2026aqua} is a recent release and shows promise with its physics-based sensor simulation, but it does not include support for generating scalable labeled data. With OceanSim 2.0, we hope to enhance our IsaacSim platform to support SDG with semantic annotations across several sensor modalities. A summary is shown in Table \ref{tab:sim_overview}. 

\begin{figure*}[t]
    \centering
    \includegraphics[width=\linewidth]{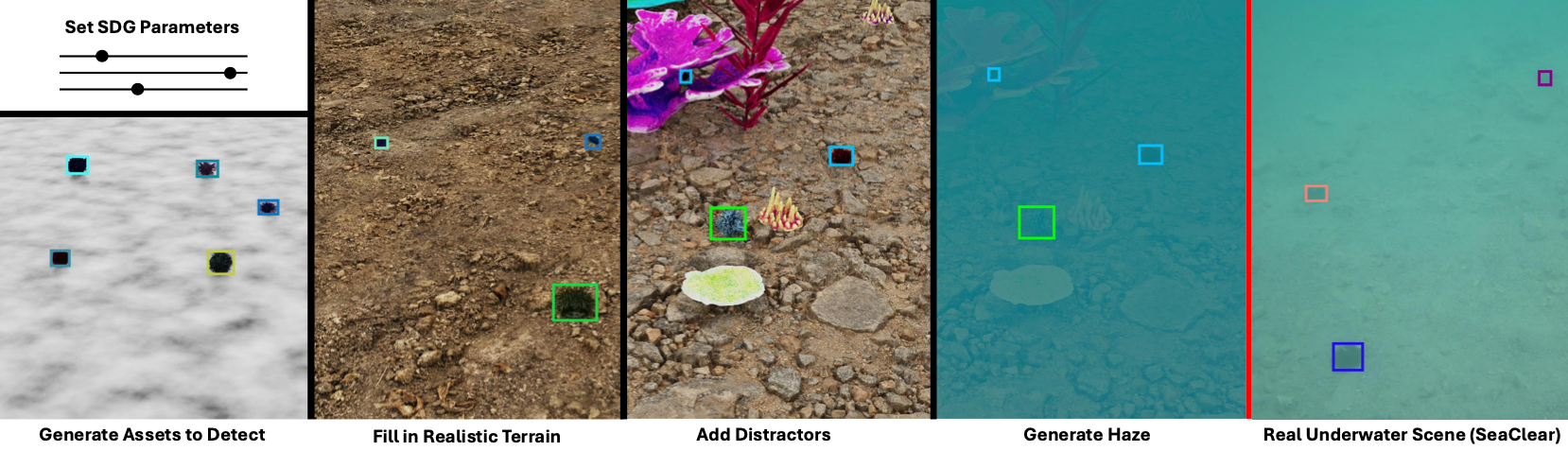}
    \caption{Our full SDG image generation pipeline compared against a real-world image from SeaClear~\cite{duras2024seaclear}. First, sea urchins are spawned on a plane mesh textured from black-white perlin noise. 
    Then, PBR textured terrain is added. Next, distractor objects are spawned.
    Finally, underwater haze effects are included. 
    Detection labels are automatically generated by renderer's underlying algorithm.}
    \label{fig:placeholder}
\end{figure*}

\subsection{Underwater Rendering}
Accurate modeling of underwater image formation is essential for generating realistic synthetic data. 
In this section, we review existing literature on modeling how light and imaging is affected in the underwater medium.

Early works in underwater rendering come from ocean optics, where the degradation in underwater images is explained through wavelength-dependent absorption and scattering~\cite{jaffe_water, monzon2024spectral}.
Jaffe et al. proposed work to decompose the effect of the underwater medium as direct transmission, forward scatter and backscatter~\cite{jaffe_water}.
Mobley et al. presented an approach where the light transporting through the water column could be modeled as a radiative transfer approach~\cite{mobley}.

To enable computationally tractable simulations and representations, researchers have adopted simplified models for the light transport through a medium.
Akkaynak et al. showed that the underwater direct transmission and backscatter terms can be modeled with different coefficients, leading to a revised model on how light radiates through water~\cite{akkaynak2018revised}.
Continued work presented Sea-thru, a method that used the geometry of the observed scene to couple the image given wavelength dependent attenuation and backscatter effects~\cite{akkaynak_sea-thru_2019}.

As part of the presented work, we leverage the model presented in Sea-thru to model what an underwater image will look like given user-defined attenuation and backscatter coefficients, along with the scene geometry.

\section{Synthetic Data Generation Pipeline}
The proposed SDG pipeline, shown in Fig.~\ref{fig:placeholder}, takes target-object meshes, distractor meshes, and scanned natural-terrain meshes as inputs. Scene composition and physical placement are procedurally randomized using the OpenUSD API and IsaacSim’s physics engine, enabling physically plausible object and camera configurations. Underwater appearance is then synthesized through deferred shading~\cite{SaitoDeferred} and the RTX renderer automatically produces synchronized ground-truth annotations during rendering. The pipeline enables generating underwater images with object labels in Kitti~\cite{Geiger2012CVPR} format. We describe each component below.

\subsection{Haze}
Underwater imaging is affected by the water medium due to light attenuation and backscattering. To simulate these effects, we implement the underwater image formation model by deferred shading following~\cite{akkaynak_sea-thru_2019}. For each color channel \( c \in \{R, G, B\} \), the observed underwater image \( I_c \) is modeled as
\begin{equation}
\label{eq:uw_model}
I_c = J_c e^{-\beta_\text{attn,c}  d} + B_{\infty,c} \left(1-e^{-\beta_\text{bs,c} d} \right),
\end{equation}
where $J$ represents the in-air image rendered using the IsaacSim Omniverse renderer, $d$ is the depth image, and $\beta_\text{attn,c}, \beta_\text{bs,c}$, and $B_{\infty,c}$ are the per channel attenuation, back-scatter, and veiling light components, respectively. The attenuation term, $ J_c e^{-\beta_\text{attn,c}  d}$, represents the decay of the original image signal as the light propagates through water, where degradation increases with increased depth, or range, from the camera, $d$. The backscatter term, $B_{\infty,c} \left(1-e^{-\beta_\text{bs,c} d}\right) $, is an additive term that accounts for the light scattered by suspended particles in water, which also increases with depth from the camera and adds a veiling luminance to the image. Although this approach is faster than volumetric rendering, enabling real-time simulation of images with underwater appearances, the trade-off is slightly less realism in the light attenuation.

\subsection{Terrain}
To replicate the realistic natural terrains, we used open-sourced real-world scanned Physically-Based Rendering (PBR) textures from PolyHaven~\cite{polyhaven} and wired textures into OmniPBR~\cite{omnipbr} materials used in IsaacSim renderer. Since our haze models rely on accurate depth computation and objects may be occluded by bumpy structures on terrain, we then deformed the base plane mesh based on the displacement map (equivalent to ``Displacement Only'' in Blender~\cite{blender_displacement}). We then pad the textures accordingly to create a sufficiently large working space to spawn objects and fill the screen.


\subsection{Objects and Distractors}
We model our simulated scenes after the SeaClear dataset, which is a real-world dataset that contains different categories of objects and creatures, with diversified morphologies~\cite{duras2024seaclear}. We focus on single-object detection in our experiments, but generating distractors and other background objects helps improve the realism of the scenes beyond just the objects we wish to detect. We generate object assets using image-to-3D generative models such as SAM3D~\cite{chen2025sam} and Hyper3D~\cite{guo2025hyper3d}. In addition to replicating target objects, we create a distractor asset pool from online asset repositories such as SketchFab~\cite{sktchfab} to increase scene diversity and robustness.

\begin{figure}[t]
    \centering
    \includegraphics[width=0.95\linewidth]{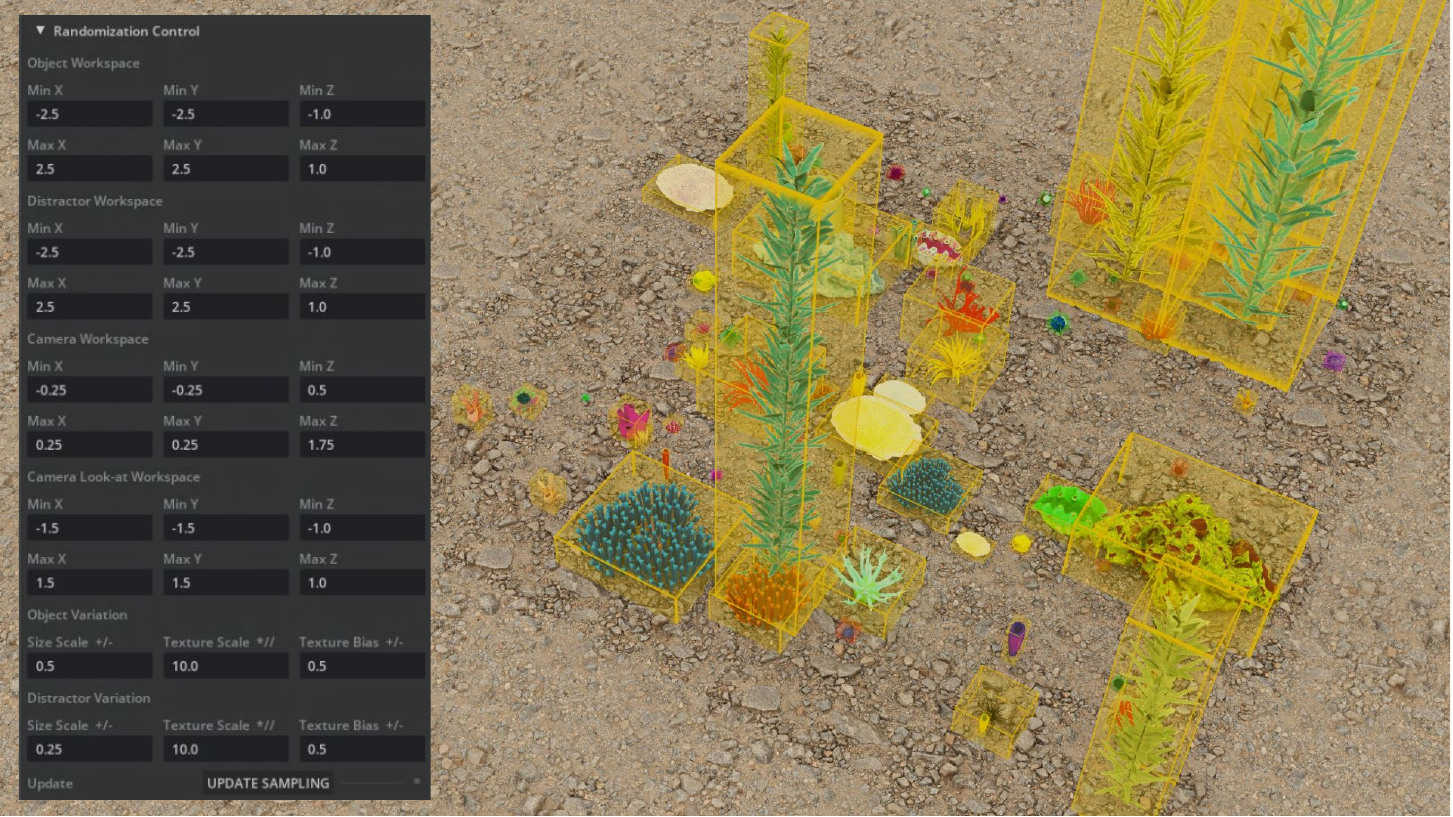}
    \caption{An SDG scene tuning playground we implemented in IsaacSim. On the left is the GUI that exposes various SDG settings including object spawning range and randomization parameters to the user. On the right is a snapshot of the native viewport with its 3D bounding box annotation turned on.}
    \label{fig:playground}
\end{figure}

\subsection{Domain Randomization}
During the generation, we have four predefined workspaces for objects, distractors, camera positions, and camera look-at vectors, respectively. Objects and distractors are sampled from vertices of the terrain mesh within the objects and distractor workspace. Camera positions are uniformly sampled in the camera workspace and look-at vectors always point toward an object within the look-at workspace. This workflow is effective in generating realistic images but requires repetitive trial and error to tune the parameters. Therefore we have made a GUI playground as shown in Fig.~\ref{fig:playground} to help users interactively find the optimal values that give the satisfactory images.

Moreover, we introduce scale and bias to the textures of the assets, which enables objects to have different colors while maintaining visual patterns. Additional randomizations include lighting variation, haze parameters, and object scale.

\section{ROS2 Integration}
Robotic research and development often makes use of the large ROS package ecosystem. We introduce integration with ROS2~\cite{ROS2}, enabling publisher-subscriber data flows between the simulator and ROS2 environments. Topic publishing is provided for all OceanSim sensors, 
as shown in Table \ref{tab:ros_topics}. 
Additionally, control commands may be sent directly to the ego robot via \texttt{cmd\_vel}.

\begin{table}[h]
    \centering
    \caption{ROS2 Topic Definitions}
    \label{tab:ros_topics}
    \begin{tabular}{ll}
        \toprule
        \textbf{Topic} & \textbf{Message Type} \\
        \midrule
        \texttt{/Barometer}               & sensor\_msg/FluidPressure \\
        \texttt{/DVL}               & geometry\_msgs/Twist \\
        \texttt{/DepthImage}              & sensor\_msgs/msg/Image \\
        \texttt{/IMU}                & sensor\_msgs/msg/Imu \\
        \texttt{/RGBCamera/image}                & sensor\_msgs/msg/Image \\
        \texttt{/ImagingSonar/image}       &  sensor\_msgs/msg/Image \\
        \texttt{/ImagingSonar/camera\_info}  & sensor\_msgs/CameraInfo \\
        \texttt{/ImagingSonar/pointcloud}    & sensor\_msgs/PointCloud2 \\
        \bottomrule
    \end{tabular}
\end{table}

\begin{table}[b]
    \centering

    \renewcommand{\arraystretch}{1.15}
    \setlength{\tabcolsep}{4.0pt}
    \footnotesize

    \caption{Ablation of synthetic scene variation on the constructed SeaClear test set~\cite{duras2024seaclear}. All models are trained with $1\times$ synthetic data, corresponding to 1,800 images.
    Best results are shown in \textbf{bold} and second-best results are \underline{underlined}. Higher is better for all metrics.}
    \begin{tabular}{@{}cccc||cccc@{}}
        \toprule
    
        \multicolumn{4}{c||}{\textbf{Synthetic Configuration}} &
        \multicolumn{4}{c}{\textbf{Real-Test Performance}} \\
    
        \cmidrule(lr){1-4}
        \cmidrule(lr){5-8}
    
        \textbf{Haze} &
        \textbf{Terr.} &
        \textbf{Distr.} &
        \textbf{Model} &
        \textbf{Prec.} &
        \textbf{Rec.} &
        \textbf{mAP}$_{50}$ &
        \textbf{mAP}$_{50\text{-}95}$ \\
        \midrule

        \multirow{2}{*}{\ding{53}} &
        \multirow{2}{*}{\ding{53}} &
        \multirow{2}{*}{\ding{53}} &
        YOLO & 0.006 & 0.024 & 0.000 & 0.000 \\
        & & & DETR & 0.237 & 0.050 & 0.036 & 0.015 \\

        \multirow{2}{*}{\checkmark} &
        \multirow{2}{*}{\ding{53}} &
        \multirow{2}{*}{\ding{53}} &
        YOLO & 0.401 & 0.180 & 0.139 & 0.053 \\
        & & & DETR & 0.227 & 0.145 & 0.113 & 0.047 \\

        \multirow{2}{*}{\ding{53}} &
        \multirow{2}{*}{\checkmark} &
        \multirow{2}{*}{\ding{53}} &
        YOLO & 0.233 & 0.024 & 0.014 & 0.004 \\
        & & & DETR & 0.401 & \textbf{0.408} & \underline{0.366} & \textbf{0.151} \\

        \multirow{2}{*}{\ding{53}} &
        \multirow{2}{*}{\ding{53}} &
        \multirow{2}{*}{\checkmark} &
        YOLO & 0.262 & 0.096 & 0.074 & 0.023 \\
        & & & DETR & \textbf{0.721} & 0.104 & 0.117 & 0.047 \\

        \multirow{2}{*}{\checkmark} &
        \multirow{2}{*}{\checkmark} &
        \multirow{2}{*}{\ding{53}} &
        YOLO & 0.334 & 0.274 & 0.148 & 0.040 \\
        & & & DETR &
        \underline{0.713} &
        \underline{0.391} &
        \textbf{0.404} &
        \underline{0.130} \\

        \multirow{2}{*}{\checkmark} &
        \multirow{2}{*}{\checkmark} &
        \multirow{2}{*}{\checkmark} &
        YOLO & 0.521 & 0.389 & 0.341 & 0.110 \\
        & & & DETR & 0.419 & 0.369 & 0.301 & 0.103 \\

        \bottomrule
    \end{tabular}
    \label{tab:scene_ablation}
\end{table}

\section{Experiments and Results}
To evaluate the sim-to-real capability of our SDG pipeline, we train a deep neural network using only synthetic images and evaluate it on real-world datasets. SeaClear is used as the real-world benchmark, as it contains water effects such as haze, and additionally has occlusion of objects of interest, and distractors~\cite{duras2024seaclear}.

\subsection{Synthetic Data Generation}
We formulate our experiment as a single-object detection task focused on sea urchins in underwater environments. 
We select sea urchins as the target because they have a relatively simple and rigid morphology, along with being a labeled object in the benchmark dataset~\cite{duras2024seaclear}. 
This makes it straightforward to collect or generate realistic 3D assets for synthetic data generation, allowing us to study the sim-to-real gap without making asset creation itself a major source of complexity. 
In practice, we use a combination of publicly available assets and assets generated with generative models.

We generate labeled synthetic data with the proposed OceanSim SDG pipeline. 
To vary both the target appearance and the surrounding environment, we use five sea urchin models, 30 natural terrain assets such as sand, river pebbles, and rock faces, and 30 distractor objects such as coral, seaweed, and clams. 
This asset diversity is used to produce a broad range of scene configurations for training and evaluation.

\subsection{Detector Training and Evaluation}
We evaluate the transfer of our synthetic data across different object detection architectures using a CNN-based detector, YOLOv9~\cite{wang2024yolov9}, and a transformer-based detector, DETR-ResNet50~\cite{carion2020end}.
We use the same image pre-processing and augmentations for both models where applicable.
All models are trained for 100 epochs, which was sufficient for convergence.

To study the effect of scene variation on sim-to-real transfer, we train models using synthetic datasets with different combinations of underwater haze, terrain diversity, and distractor objects, as shown in Table~\ref{tab:scene_ablation}.
We also vary the amount of synthetic training data to evaluate the effect of dataset size, shown in Table~\ref{tab:size_ablation}.

We evaluate all models on the same real-world test set of 179 images containing 718 annotated sea urchins.
The test set is a representative sample of SeaClear~\cite{duras2024seaclear}, with varying underwater conditions, urchin size, and urchin spatial distribution.
We report $\mathrm{mAP}_{50}$ and $\mathrm{mAP}_{50\text{-}95}$, together with precision and recall at the operating point that maximizes the F1 score.

For comparison, we train both architectures on real-world data using the same training-set size as the $1\times$ synthetic configuration.
Their performance, reported in Table~\ref{table:baseline}, provides a real-data reference for evaluating the sim-to-real gap.

\begin{table}[t]
    \centering
    
    \caption{Effect of synthetic dataset size on real-world detection performance.
    Haze, terrain variation, and distractor objects are enabled for all
    configurations. A dataset size of $1\times$ corresponds to 1,800 images.
    Best results are shown in \textbf{bold} and second-best results are \underline{underlined}. Higher is better for all metrics.}
    \label{tab:size_ablation}

    \renewcommand{\arraystretch}{1.15}
    \setlength{\tabcolsep}{8.0pt}
    \footnotesize

    \begin{tabular}{@{}cccccc@{}}
        \toprule
        \textbf{Size} &
        \textbf{Model} &
        \textbf{Prec.} &
        \textbf{Rec.} &
        \textbf{mAP}$_{50}$ &
        \textbf{mAP}$_{50\text{-}95}$ \\
        \midrule

        \multirow{2}{*}{$1\times$}
        & YOLO & 0.521 & 0.389 & \underline{0.341} & \underline{0.110} \\
        & DETR & 0.419 & 0.369 & 0.301 & 0.103 \\
        \addlinespace[1.5pt]

        \multirow{2}{*}{$5\times$}
        & YOLO & \textbf{0.557} & 0.372 & 0.284 & 0.103 \\
        & DETR & 0.519 & \textbf{0.521} & \textbf{0.409} & \textbf{0.168} \\
        \addlinespace[1.5pt]

        \multirow{2}{*}{$10\times$}
        & YOLO & \underline{0.544} & \underline{0.442} & 0.327 & 0.109 \\
        & DETR & 0.392 & 0.290 & 0.180 & 0.075 \\

        \bottomrule
    \end{tabular}
\end{table}

\begin{figure*}[t]
    \centering
    \includegraphics[width=\linewidth]{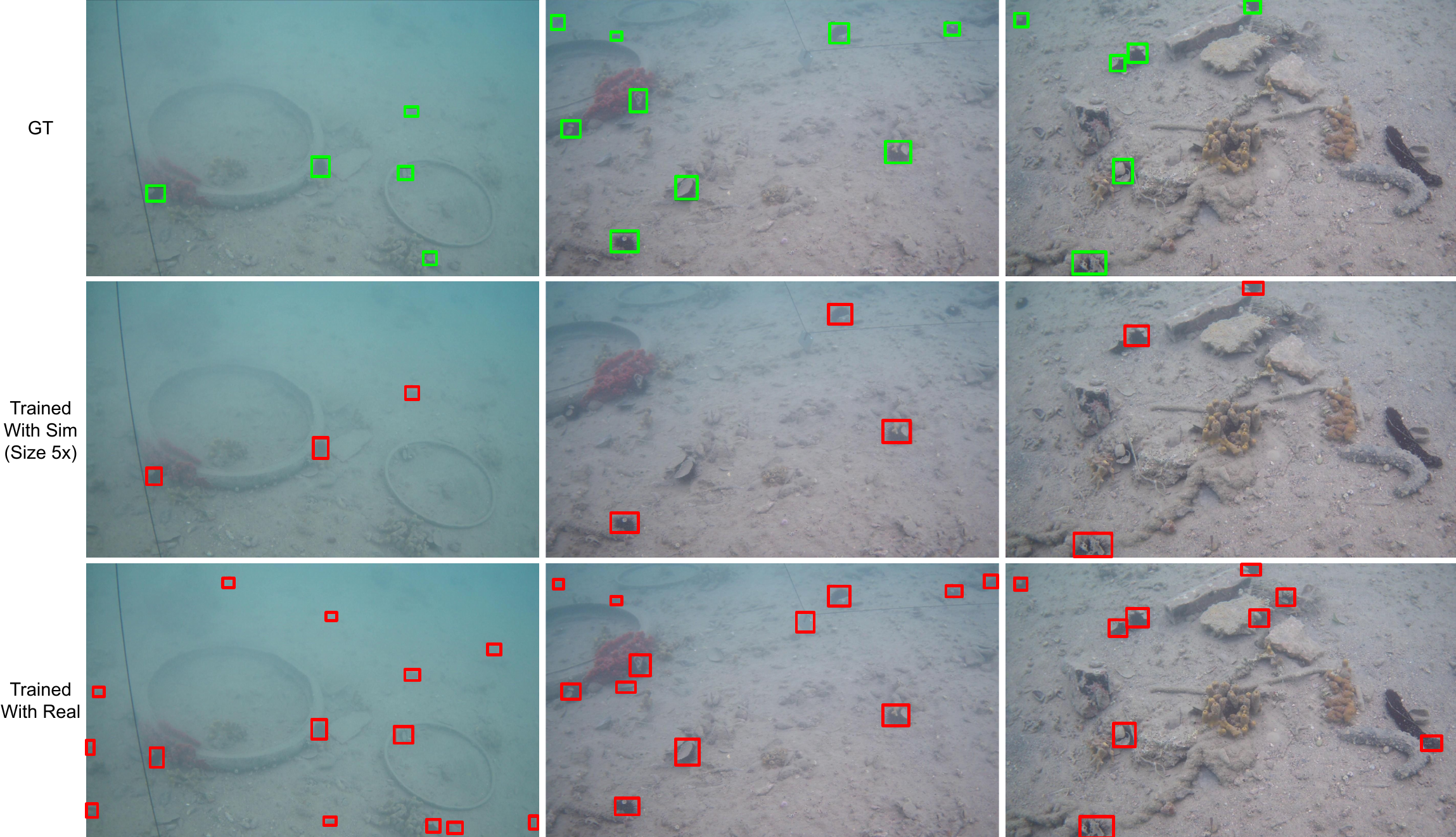}
    \caption{Qualitative comparison of the best-performing models trained on synthetic and real-world data. 
    The first row shows the ground-truth labels from the SeaClear dataset~\cite{duras2024seaclear}. 
    The second row shows predictions from the DETR trained with the $5\times$ simulated dataset. The third row shows predictions from the DETR model trained on the real-world training dataset, as reported in Table~\ref{table:baseline}. 
    }
    \label{fig:qualitative}
\end{figure*}
\subsection{Sim-to-Real Performance}
The results from Table~\ref{tab:scene_ablation} indicate that each of the three forms of scene variation can improve sim-to-real performance, although their effects differ across the two detector architectures.

Without simulated haze, the synthetic images do not reproduce the changes in object appearance caused by attenuation and backscatter in real underwater imagery. 
The resulting model therefore produces more false negatives when sea urchins are weakly visible or partially blended with the background.

DETR tends to learn more information with a diversified background (terrain) than haze. This may arise from the fact that CNNs have hard-coded inductive bias while transformers will learn the bias~\cite{cordonnier2021cnn_transformer}. Moreover, YOLO focuses on local features therefore emphasizing the alignment of pixel appearance between simulated and real data, while DETR better learns the spatial structure from the image including terrains and occlusions~\cite{wang2018nonlocal}.

Distractors can serve as a domain randomization technique to increase the generalizability to varying scene structure; therefore, introducing distractors can only be meaningful with augmented haze and terrains otherwise the domain shift is too large from the real-world. By looking at mAP alone, the DETR transformer model generally performs better than the YOLOv9 CNN, but does not benefit more from increasing the training size~\cite{dosovitskiy2021vit}, which implies the generated images need to cover more domain distributions rather than increased dataset sizes. 

Comparing to models trained with real-world data (Fig.~\ref{fig:qualitative}, Table~\ref{table:baseline}), models trained on simulated data from our SDG pipeline show lower recall. Models trained on simulated data fall short on being able to identify objects that are under heavy underwater degradation. For example, sea urchins that are far from the camera and are thus largely blurred by backscatter and attenuation are challenging to detect for models trained on simulated data. Another challenge is that the real-world training dataset is captured from the same underwater region, making it very close to the test distribution. Although this leads to higher performance on the SeaClear test set, we posit that the generalizability of the model trained on the real data is much weaker.

\begin{table}[t]
    \renewcommand{\arraystretch}{1.15}
    \setlength{\tabcolsep}{8.0pt}
    \footnotesize
    \centering
    \caption{Baseline model performance when trained on the training set of the real dataset from SeaClear, composed of 1800 images.}
    \begin{tabular}{l c cccc}
        \toprule
        \textbf{Model} && \textbf{Prec.} & \textbf{Rec.} & \textbf{mAP$_{50}$} & \textbf{mAP$_{50-95}$} \\
        \midrule
        YOLO && 0.821 & 0.724 & 0.793 & 0.403 \\
        DETR && 0.816 & 0.804 & 0.838 & 0.347 \\
        \bottomrule
    \end{tabular}
    \label{table:baseline}
\end{table}

\section{Discussion}
Overall, our experiments show that synthetic data generated with OceanSim can transfer to real underwater imagery, but a substantial sim-to-real gap remains.
Increasing the amount of synthetic data does not consistently improve performance, which suggests that the main limitation involves factors beyond the number of generated images or the training dataset size.
Instead, we believe that our results indicate that the fidelity and diversity of the data are the main limitations of the current pipeline.

First, a prominent limitation is the simplified representation of underwater appearance.
Our current rendering model captures effects such as attenuation and backscatter, but does not fully model the spatially varying and view-dependent nature of underwater light transport~\cite{MulticolorZhou}.
Other common visual effects, such as caustics, suspended particles, and marine snow, which have been shown to help with closing the simulation to real gap, are also not represented in detail~\cite{bagoren2026surfslam}.
Modeling these effects at scene scale can be computationally expensive, and the closed-source IsaacSim RTX renderer further limits modifications to the underlying light-transport model.
Future versions of SDG for underwater image generation may therefore benefit from improved appearance modeling.

Scene diversity is another important limitation.
Although we constrained our detection problem to encompass a simple geometry and we randomized the scene through multiple factors, the number of underlying assets remains limited.
Generating additional images from similar assets can therefore produce many samples with similar visual and geometric structure, meaning that an increased number of images does not necessarily capture a larger distribution of information.
Techniques that enable more scalable scene generation, which involves terrain and asset generation, could provide broader domain coverage, continuing in the direction presented in works such as Zhang et al.~\cite{zhang2025infiniteleaguesseaphotorealistic}.

Finally, evaluating generalization itself remains difficult.
There is no sufficiently diverse underwater object-detection benchmark.
In addition to the challenges of collection of underwater data in the field, the curation of existing datasets for machine learning models remains just as critical, but is an arduous task to set as a benchmark for the field.
The real-world test set we leveraged for experiments in this paper still represents a limited set of environments and imaging conditions, and therefore cannot determine whether improvements generalize across different sites, water conditions, illumination, terrain, and sensor configurations.
We therefore view these experiments primarily as a study of the current sim-to-real limitations rather than as evidence of general sim-to-real performance.
A broader and more diverse real-world benchmark would be necessary for a stronger evaluation of generalization.

\section{Conclusion}
Although our synthetic data cannot fully recover the real-world distributions, the pipeline is able to generate useful underwater RGB images that can be used to train zero-shot networks to perform object detection. The entire workflow only requires a few real-world images for generative networks to replicate the assets. 
Based on the low recall score of models trained on simulated data compared to models trained on real-world data, one of the most prominent future improvements is a more sophisticated light transport formulation that can better align the hazy appearance of objects with the real-world. 
We also hope that this study creates interest in better real-world object detection benchmarks, which will aid in the quantitative assessment of various underwater simulators for perceptual tasks.



\bibliographystyle{IEEEtran}
\bibliography{references}
\end{document}